\documentclass[11pt,a4paper]{article}
\usepackage[hyperref]{emnlp-ijcnlp-2019}
\aclfinalcopy

\usepackage{slashbox} 

\usepackage{CJKutf8}

\usepackage{hyperref}
\usepackage{url} 

\usepackage{times}
\usepackage{latexsym}
\usepackage{xcolor}
\usepackage{bm}

\usepackage{url}

\usepackage{tikz}
\usepackage{tikz-qtree}
\usepackage{latexsym}
\usepackage{graphicx}			
\usepackage{times}
\usepackage{latexsym}
\usepackage{mathpartir}
\usepackage{pifont}

\usepackage{proof}
\usepackage{xspace}
\usepackage{multirow}
\usepackage{forest}
\usepackage{subcaption}
\usepackage{multirow}
\usepackage{CJKutf8}
\usepackage{tabu} 
\usepackage{subcaption}

\usepackage{dsfont}
\usepackage{amsmath}
\usepackage{amssymb}
\usepackage{amsthm}
\usepackage{algorithm}
\newcommand{\namecite}[1]{\newcite{#1}}

\newcommand{\tuple}[1]{\ensuremath{\langle {#1} \rangle}}

\newcommand{\notes}[1]{}

\theoremstyle{definition}

\theoremstyle{plain}

\newcommand{\vech}{\ensuremath{\bm{h}}}

\newcommand{\ith}[1]{\ensuremath{i^{{th}}}}

\newcommand{\chn}[1]{\mbox{{\it {#1}}}}

\newcount\permx
\newcount\permy
\def\permdot#1#2{
\permx=#1 \advance\permx by-1
\permy=#2 \advance\permy by-1
\psframe[fillcolor=black, fillstyle=solid]
(\permx,\permy)(#1, #2)
}

\newcommand{\toptop}{\operatornamewithlimits{\mathrm{top}}}
\newcommand{\nextfunc}{\operatornamewithlimits{\mathrm{next}}}

\newcommand{\startsym}{\mbox{\scriptsize \texttt{<s>}}\xspace}

\newcommand{\boxnum}[1]{{\setlength{\fboxsep}{1pt}\raisebox{1pt}{\hspace{1pt}\fbox{\tiny #1}\hspace{1pt}}}}
\newcommand{\ind}[1]{\ensuremath{_{\kern-0.5pt\boxnum{#1}}}}

\newcommand{\vecx}{\ensuremath{\bm{x}}\xspace}
\newcommand{\vecy}{\ensuremath{\bm{y}}\xspace}

\def\namecite{\newcite}

\newcommand{\smallnt}[1]{\ensuremath{_{\mbox{\tiny PP}}}\xspace}

\newcommand{\pseudocode}{Algorithm}
\floatname{algorithm}{\pseudocode}

\iffalse

\else

\fi

\newcommand{\eos}{\mbox{\scriptsize \texttt{<eos>}}\xspace}

\newcommand{\waitk}{\ensuremath{\text{wait-$k$}}\xspace}

\definecolor{chocolate}{rgb}{0.28, 0.02, 0.03}
\definecolor{PaleGreen}{rgb}{0.33, 0.545,0.33}
\definecolor{colorC0}{RGB}{51,113, 169}
\definecolor{colorC1}{RGB}{243,130,37}
\usepackage{array}
\usepackage{diagbox}
\usepackage{balance}

\title{Speculative Beam Search for Simultaneous Translation 
} 

\author{Renjie Zheng $^{2,}$\thanks{\; These authors contributed equally.} \,
  Mingbo Ma $^{1, \ast}$ \,
  Baigong Zheng $^{1}$ \,
  Liang Huang $^{1,2}$
\\
  $^{1}$Baidu Research, Sunnyvale, CA, USA \\
  $^{2}$Oregon State University, Corvallis, OR, USA \\
  \texttt{zrenj11@gmail.com } \\
  \texttt{\{mingboma, baigongzheng, lianghuang\}@baidu.com } \\
}

\date{}

\begin{document}
\begin{CJK}{UTF8}{gbsn}
\maketitle
\begin{abstract}

Beam search is universally used in full-sentence translation 
but its application to simultaneous translation remains non-trivial,
where output words are committed on the fly.
In particular, the recently proposed \waitk policy \cite{ma+:2018}
is a simple and effective method that (after an initial wait)
commits one output word on receiving each input word, making beam search seemingly impossible.
To address this challenge,
we propose a speculative beam search algorithm that hallucinates several steps into the future in order to reach a more accurate decision,
implicitly benefiting from a target language model. 
This makes beam search applicable for the first time to the generation of a single word in each step.
Experiments over diverse language pairs show large improvements over previous work. 

\end{abstract}

\section{Introduction}

Beam search has been widely used in neural text 
generation such as machine translation
\cite{sutskever+:2014,bahdanau+:2014a},
summarization \cite{rush+:2015,ranzato+:2016},
and image captioning \cite{vinyals+:2015,xu+:2015}.
It often leads to substantial improvement 
over greedy search
and
becomes an essential component in almost all text generation systems.

However, beam search is easy for the above tasks because they are all
{\em full-sequence} problems, where the whole input sequence is available at the beginning
and the output sequence only needs to be revealed in full at the end.
By contrast, in language and speech processing,
there are many {\em incremental processing} tasks with {\em simultaneity requirements},
where the output needs to be revealed to the user incrementally without revision (word by word, or in chunks)
and the input is also being received  incrementally.
Two most salient examples are streaming speech recognition \cite{chiu+:2018},
widely used in speech input and dialog systems (such as Siri),
and simultaneous translation \cite{bangalore+:2012,oda+:2015,grissom+:2014,jaitly+:2016}, widely used in
international conferences and negotiations.
In these tasks, the use of full-sentence beam search becomes
seemingly impossible as output words need to be committed on the fly.

How to adapt beam search for such incremental tasks in order to improve their generation quality?
We propose a general technique of {\em speculative beam search} (SBS),
and apply it to simultaneous translation.
At a very high level, 
to generate a single word,
instead of simply choosing the highest-scoring one (as in greedy search),
we further speculate $w$ steps into the future, and use
the ranking at step $w\!+\!1$ to reach a more informed decision for step 1 (the current step);
this method implicitly benefits from a target language model,
alleviating the label bias problem in neural generation \cite{murray+chiang:2018,ma2019learning}.

We apply this algorithm to two representative approaches to simultaneous translation:
the fixed policy method \cite{ma+:2018} and the
adaptive policy method \cite{gu+:2017}.
In both cases, we show that SBS improves translation quality while
maintaining latency (i.e., simultaneity).

\if
Recently, there have been two main approaches to simultaneous translation

In the application of simultaneous translation \cite{ma+:2018,gu+:2017},
whose source side is gradually revealed and 
target side are committed incrementally,
the system needs to generate {\em{immediate}}, {\em{inalterable}} 
outputs on the fly.
Especially for wait-$k$ model \cite{ma+:2018},
which commits one output word on receiving each input word,
conventional beam search becomes inapplicable.

\fi


\if
In conventional beam search,
the decoder buffers top $b$ candidates based on their cumulative 
model score at time step $t$, 
and advances to $t+1$ step given
the previously buffered candidates from $t$ as the predecessors.
The above generation continues until
beam search satisfies the optimal 
stopping criteria \cite{huang+:2017}.\footnote{Beam search 
terminates when the score of the top unfinished hypothesis 
is worse than any finished hypothesis, or the
\eos is the most probable candidate in the beam at some step.};
or $t$ reaches the maximum, predefined length limits (i.e., 100).
When beam search terminates, the best candidate will be selected
at the last step, and decoder backtracks its ancestor path
to reveal the previous words.
However, as we notice, 
all the words of the intermediate steps will not be committed
until the beam selects the best candidate at the last step.
This limits the beam search usage scenario
Thus, we need a novel beam search technique, 
which can be applied to 
the scenarios which enables {\em{immediate}}, {\em{inalterable}} 
outputs on the fly.
\fi

\if
We therefore propose a novel and simple Speculative Beam Search (SBS) 
which will alway speculate the future target contents by 
operating the beam search with small amount of steps forward.
In the case of wait-$k$ framework\cite{ma+:2018}, which
only perform greedy search on one word during intermediate steps,
we implement a beam search on each word with several steps forward,
and return the ancestor who has better offspring instead of 
greedy choice.
Our proposed SBS is a general technique that can be easily applied to
any other types of simultaneous translation framework.
Although in the original RL-based simultaneous translation
system with adaptive policy \cite{gu+:2017} which already employed
conventional beam search on small chunks of words,
our propose SBS still be can applied on top of it and 
boost the performance even further.
\fi


\section{Preliminaries}

We first review standard full-sentence NMT and 
beam search to set up the notations,
and then review different approaches to simultaneous MT.

\subsection{Full Sentence NMT and Beam Search}
\begin{figure*}[!ht]
\centering
\includegraphics[width=16.cm]{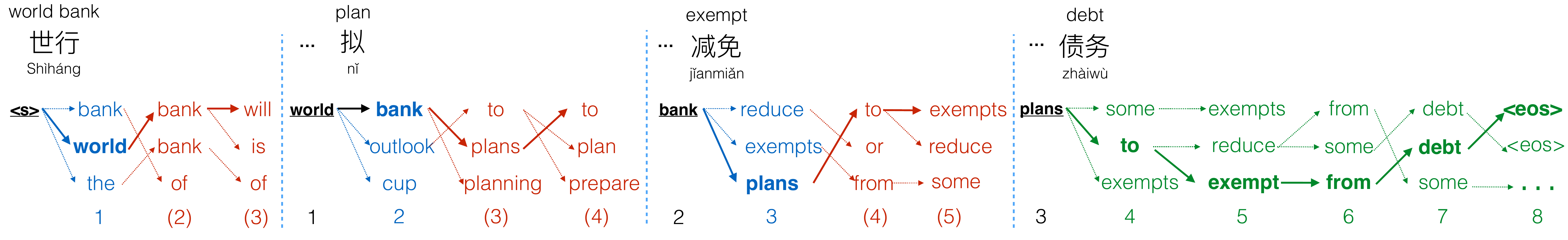}
\captionof{figure}{Wait-$1$ policy example 
to illustrate the procedure of 
SBS. The top Chinese words are the source side inputs
which are incrementally revealed to the encoder. Gloss is annotated above
Chinese word and Pinyin is underneath.
There are two extra steps (speculative window) 
are taken (red part) beyond greedy.
When source reaches the last word ``债务'' (debt), 
the decoder gets into tail and
performs conventional beam search (in green).
}
\label{fig:specmodle}
\vspace{-0.5cm}
\end{figure*}

The encoder processes the input sequence 
$\vecx = (x_1,...,x_n)$, where $x_i \in \mathbb{R}^{d}$ represents 
an input token as a $d$ dimensional vector,
and produces a new list of hidden states 
$\vech =f(\vecx) = (h_1,...,h_n)$ to represent  \vecx.
The encoding function $f$ 
can be 
RNN, CNN or Transformer. 

On the other hand,
the (greedy) decoder  
selects the highest-scoring word $y_t$ given 
source representation $\vech$ and previously 
generated target tokens, $\vecy_{<t}=(y_1,...,y_{t-1})$.
The greedy search continues 
until it emits \eos,
and the final hypothesis $\vecy = (y_1,...,y_t)$ with $y_t = \eos$ 
\vspace{-0.1cm}
\begin{equation}
\vspace{-0.1cm}
p(\vecy \mid \vecx) = \textstyle\prod_{t=1}^{|\vecy|}  p(y_t \mid \vecx,\, \vecy_{<t})
\label{eq:gensentscore}
\end{equation}



As greedy search only explores one single path among 
exponential many alternatives, 
beam search is used to improve the search. 
At each step~$t$, it maintains a beam $B_t$ of size $b$, which is an ordered list of $\tuple{\text{hypothesis, probability}}$ pairs;
for example $B_0 = [\tuple{\startsym, 1}]$.
We then define one-step transition from the previous beam to the next as
\begin{equation}
 \textstyle \nextfunc_1^b(B)\! = \!\toptop^b  
  \{\tuple{\vecy\!\circ v, \ s\!\cdot\! p(v | \vecx, \vecy)} \mid \tuple{\vecy\!, s} \!\in\! B \} \notag
\end{equation}
where $\toptop^b (\cdot)$ returns the top-scoring $b$ pairs,
and $\circ$ is the string concatenation operator.
Now $B_t = \nextfunc_1^b(B_{t-1})$.
As a shorthand, we also define the multi-step beam search function recursively:
\begin{equation}
\vspace{-0.1cm}
 \textstyle \nextfunc^b_i(B)\! = \!\nextfunc^b_{1}(\nextfunc^b_{i-1}(B)) 
\end{equation}
Full-sentence beam search (over a maximum of $T$ steps) 
yields the best hypothesis $\vecy^*$ with score $s^*$
(see \namecite{huang+:2017} for stopping criteria):
\begin{equation}
\vspace{-0.1cm}
 \textstyle \tuple{\vecy^*, s^*} = \!\toptop^1 (\nextfunc^b_T([\tuple{\startsym,1}]))
\end{equation}

\if
As we notice from above definition of conventional beam search,
the approximate best candidate only can be selected 
when beam search terminates which is at the end of decoding step.
However, in some applications such as 
simultaneous machine translation \cite{ma+:2018,gu+:2017},
which requires 
{\em{immediate}}, {\em{inalterable}} 
outputs at each step or some certain steps 
which are much earlier than the final step,
there is no simple adaptations or 
even inapplicable with 
the conventional beam search.
\fi

\subsection{Simultaneous MT: Policies and Models}

There are two main categories of policies in 
neural simultaneous translation decoding (Tab.~\ref{tb:3methods}):

\begin{enumerate}
\item
  The first method is to use 
  a fixed-latency policy, such as the \waitk policy \cite{ma+:2018}.
  Such a method would, after an initial wait of $k$ source words,
  commit one target word on receiving each new source word.
  When the  source sentence ends, the decoder can
  do a {\em tail beam search} on the remaining target words,
  but beam search is seemingly impossible before the source sentence ends.
\item
  The second method learns an adaptive policy
  which uses either supervised  \cite{zheng2019simpler}
or reinforcement learning  \cite{grissom+:2014,gu+:2017}
to decide whether to READ (the next source word) or WRITE (the next target word) . 
Here the decoder can commit a chunk of  
{\em{multiple words}} for a series of consecutive WRITEs.
\end{enumerate}

In terms of modeling (which is orthogonal to decoding policies),
we can also divide most simultaneous translatoin efforts into two camps:

\begin{enumerate}
\item Use the standard full-sentence translation model trained by classical seq-to-seq \cite{dalvi+:2018,gu+:2017,zheng2019simpler}.
  For example, the ``test-time wait-$k$'' scheme \cite{ma+:2018} uses
  the full-sentence translation model and performs \waitk decoding at test time.
  However, the obvious training-testing mismatch in this scheme usually leads to inferior quality.
\item Use a genuinely simultaneous model trained by the recently proposed prefix-to-prefix framework \cite{ma+:2018,arivazhagan2019monotonic,zheng2019simultaneous}.
  There is no training-testing mismatch in this new scheme, with the cost of slower training.
\end{enumerate}

\newcommand{\smallcite}[1]{{\small \cite{#1}}}

\begin{table}
  \resizebox{.49\textwidth}{!}{
\begin{tabular}{|p{1.2cm}||p{3.7cm}|p{3.35cm}|}
  \hline
   \multirow{2}{*}{\hspace{-0.05cm}\backslashbox[1.7cm]{policy}{model}}    & \em sequence-to-sequence       & \em prefix-to-prefix \\
                          & (full-sentence model) & (simultaneous model)\\
  \hline
\em fixed-latency    & test-time \waitk\ \smallcite{dalvi+:2018,ma+:2018}   & \waitk\ \smallcite{ma+:2018}\\
\hline
\multirow{4}{*}{\em adaptive}& RL  &   MILk \\
&    \smallcite{gu+:2017} &   \smallcite{arivazhagan2019monotonic}  \\
  & Supervised Learning &  Imitation Learning \\
              &  \smallcite{zheng2019simpler} & \smallcite{zheng2019simultaneous}\\
\hline
\end{tabular}
}
\captionof{table}{Recent advances in simultaneous translation.}
\label{tb:3methods}
\vspace{-0.6cm}
\end{table}

\section{Speculative Beam Search}


We first present our speculative beam search 
on the fixed-latency wait-$k$ policy (generating a single word per step),
and then adapt it
to the adaptive policies (generating multiple words per step).


\subsection{Single-Step SBS}

The \waitk policy conducts translation concurrently with the source input,
committing output words one by one 
while the source sentence is still growing.
In this case, conventional beam search is clearly inapplicable.

\begin{figure}[!bt]
\centering
\includegraphics[width=7.cm]{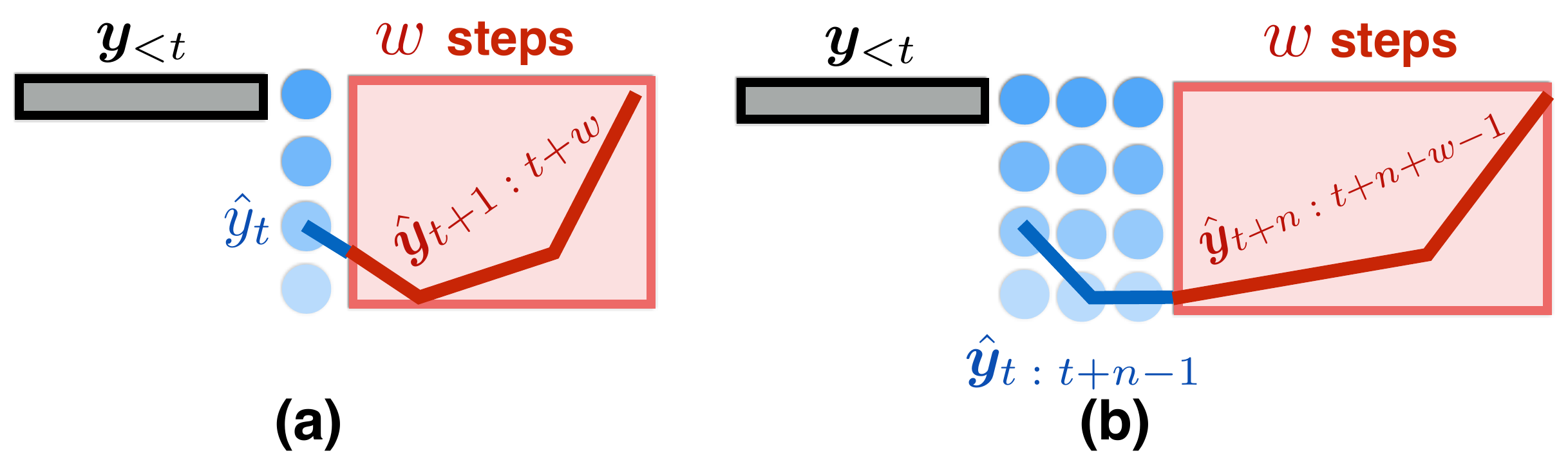}
\captionof{figure}{
Illustration of SBS:
(a) wait-$k$ policy (Eqs.~\ref{eq:specwaitk1}--\ref{eq:specwaitk});
(b) adaptive policy (Eqs.~\ref{eq:specchunk1}--\ref{eq:specchunk}).
Speculations in red.
}
\label{fig:beam}
\vspace{-0.5cm}
\end{figure}

We propose to perform
speculative beam search at each step by hallucinating $w$ more steps into the future,
and use the ranking after these $w+1$ steps to make a more informed decision for the current step.
More formally, at step $t$, we generate $y_t$ based on already committed prefix $\vecy_{<t}$:
\begin{align}
\textstyle \tuple{\hat{\vecy},s_t} &= \textstyle{\toptop^1} \bigl(\nextfunc^b_{1+w}([\tuple{\vecy_{<t},1}])\bigr)
\label{eq:specwaitk1}\\
\textstyle {\vecy}_{\leq t} &= \vecy_{<t} \circ \hat{y}_t
\label{eq:specwaitk}
\vspace{-0.2cm}
\end{align}
where $\hat{\vecy} = \vecy_{< t}\circ \hat{y}_t \circ \hat{\vecy}_{t+1:t+w}$
has three parts, with the last one being a speculation
of  $w$ steps (see Fig.~\ref{fig:beam}).
We use $\nextfunc^b_{1+w}(\cdot)$ 
to speculate  $w$ steps.
The candidate $\hat{y}_t$ is selected 
based on the accumulative model score $w$ steps later.
Then we commit $\hat{y}_t$ and move on to step $t+1$.


In the running example  in Fig.~\ref{fig:specmodle}, 
we have $w=2$ and $b=3$. 
In the greedy mode, after the wait-$1$ policy receives the first source
word, ``世行'' (world bank), 
the basic wait-$1$ model commits ``bank'' which has 
the highest score.
In SBS, we perform a  beam search for $1+w=3$ steps 
with the two speculative steps marked in red.		
After 3 steps, 
the path ``world  bank will''
becomes the top candidate,
thus we choose to commit ``world'' instead of ``bank''  and
restart a new speculative beam search with ``world''
when we receive a new source word, ``拟''(plan to);
the speculative part from the previous step (in red) is removed.

\subsection{Chunk-based SBS}

The RL-based adaptive policy system~\cite{gu+:2017}
can commit a chunk of multiple words whenever there is a series of consecutive WRITEs,
and conventional beam search can be applied on each chunk
to improve the search quality within that chunk, which is already used in that work.


However, on top of the obvious per-chunk beam search,
we can still apply SBS to
further speculate $w$ steps after the chunk.
For a chunk of length $n$ starting at position $t$, we adapt SBS as:
\begin{align}
\textstyle \tuple{\hat{\vecy},s_t} &= \textstyle{\toptop^1} \bigl(\nextfunc^b_{n+w}([\tuple{\vecy_{<t},1}])\bigr)
\label{eq:specchunk1}\\
\textstyle {\vecy}_{\leq t+n-1} &= \vecy_{<t} \circ \hat{\vecy}_{t:t+n-1}
\label{eq:specchunk}
\end{align}
Here $\nextfunc_{n+w}^b(\cdot)$ does a beam search of $n+w$ steps, with the last $w$ steps speculated.
Similarly,
\[\hat{\vecy} = \vecy_{<t}\circ \hat{\vecy}_{t:t+n-1} \circ \hat{\vecy}_{t+n:t+n+w-1}\]
has three parts, with the last being a speculation
of $w$ steps, and the middle one being the chunk of $n$ steps returned and committed
(see Fig.~\ref{fig:beam}).



\section{Experiments}

\subsection{Datasets and Latency Metrics}

\begin{figure*}[htb]
\begin{tabular}{cc}
\centering
  \includegraphics[width=6.9cm]{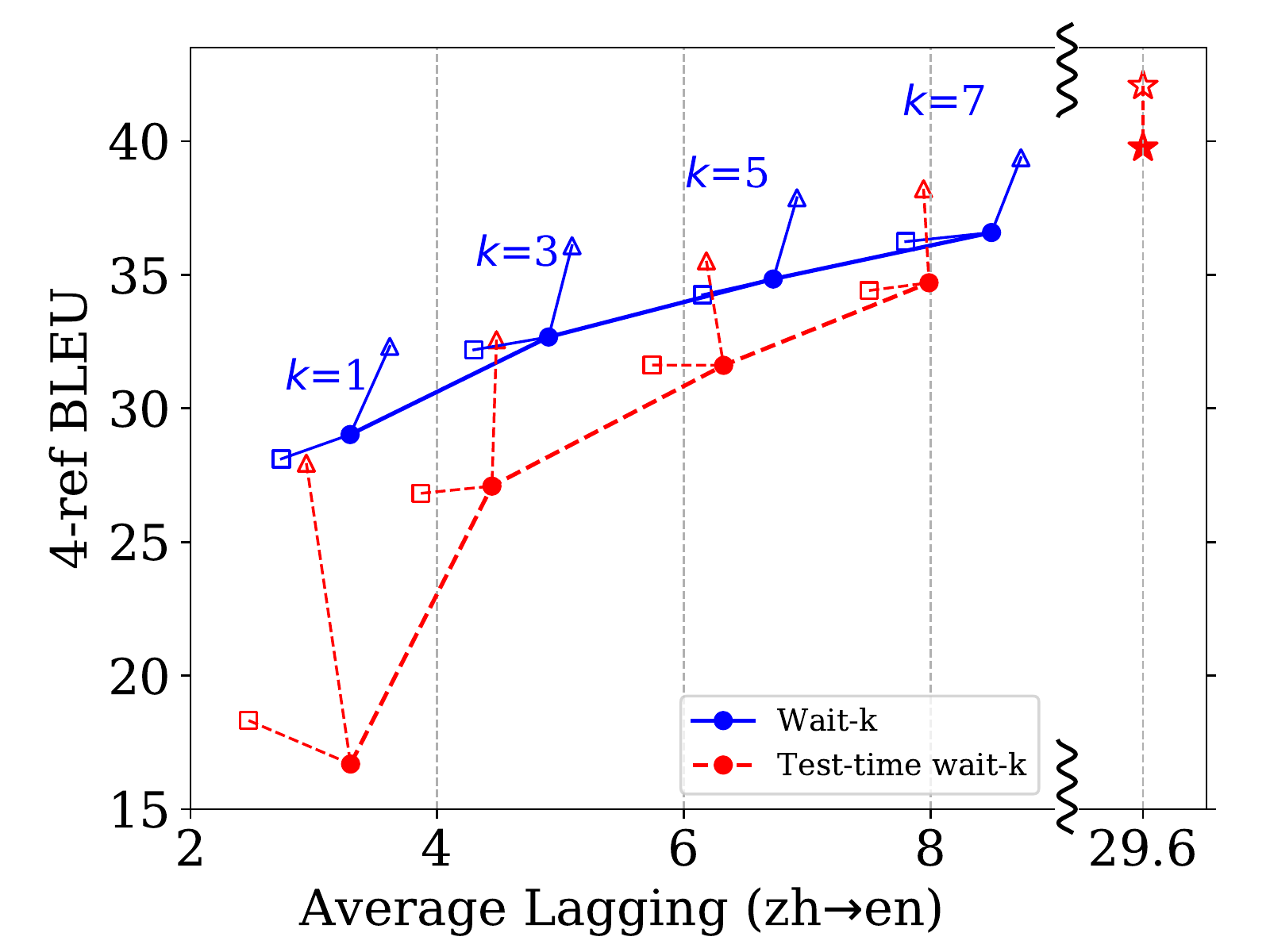} & \qquad
\centering
  \includegraphics[width=6.9cm]{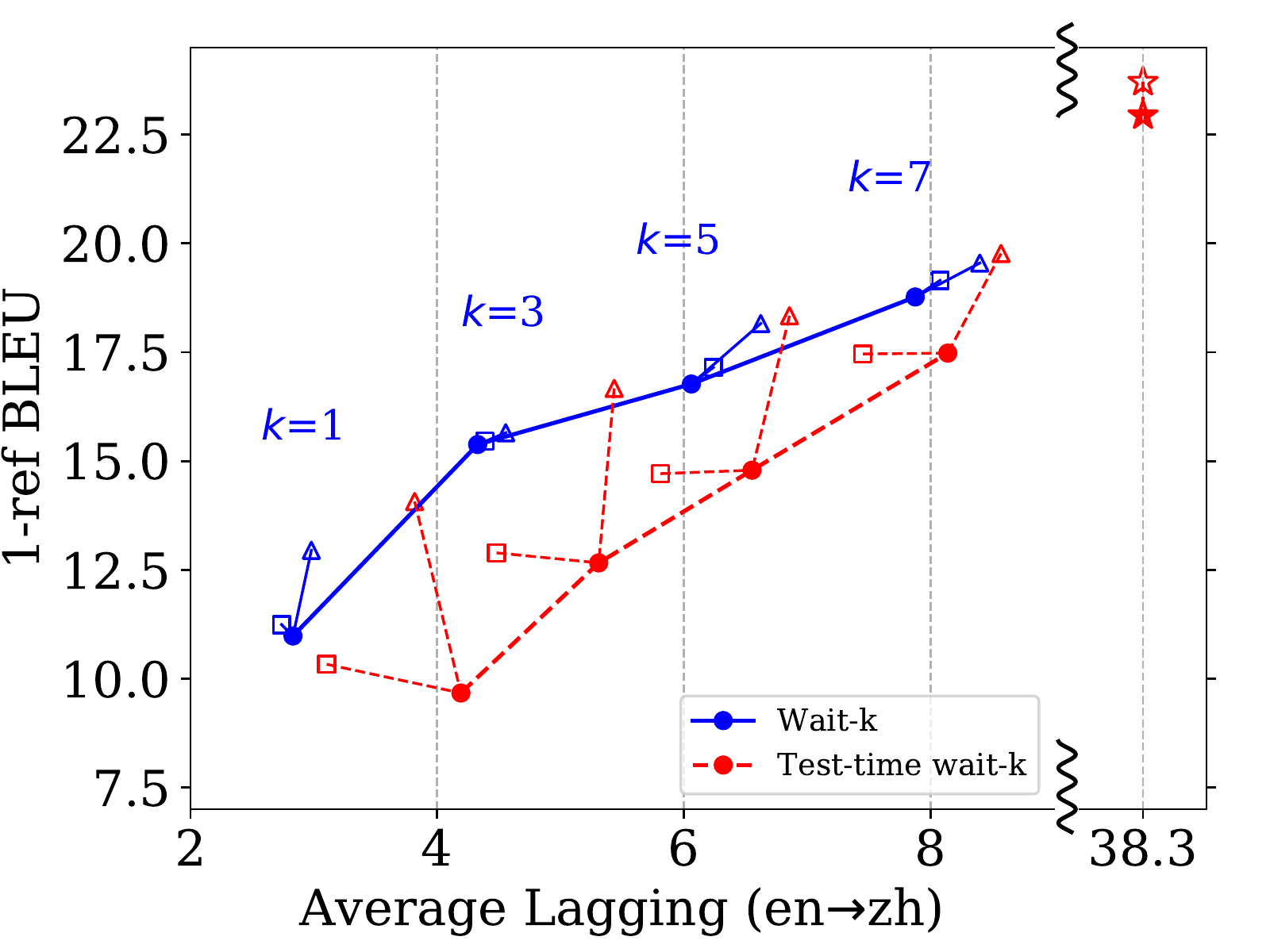}
\end{tabular}\\[-0.4cm]
\captionof{figure}{
  BLEU against AL 
  using wait-$k$ model.
  \textcolor{red}{$\square$} \textcolor{blue}{$\square$}: conventional beam search only in target tail (when source finishes).
  \textcolor{red}{$\triangle$} \textcolor{blue}{$\triangle$}: speculative beam search.
  \textcolor{red}{$\bigstar$\ding{73}}:full-sentence baseline (greedy and beam-search).
}
\label{fig:zh2en_al}
\end{figure*}

\begin{figure*}[tb] 
\begin{tabular}{cc}
\centering
  \includegraphics[width=6.9cm]{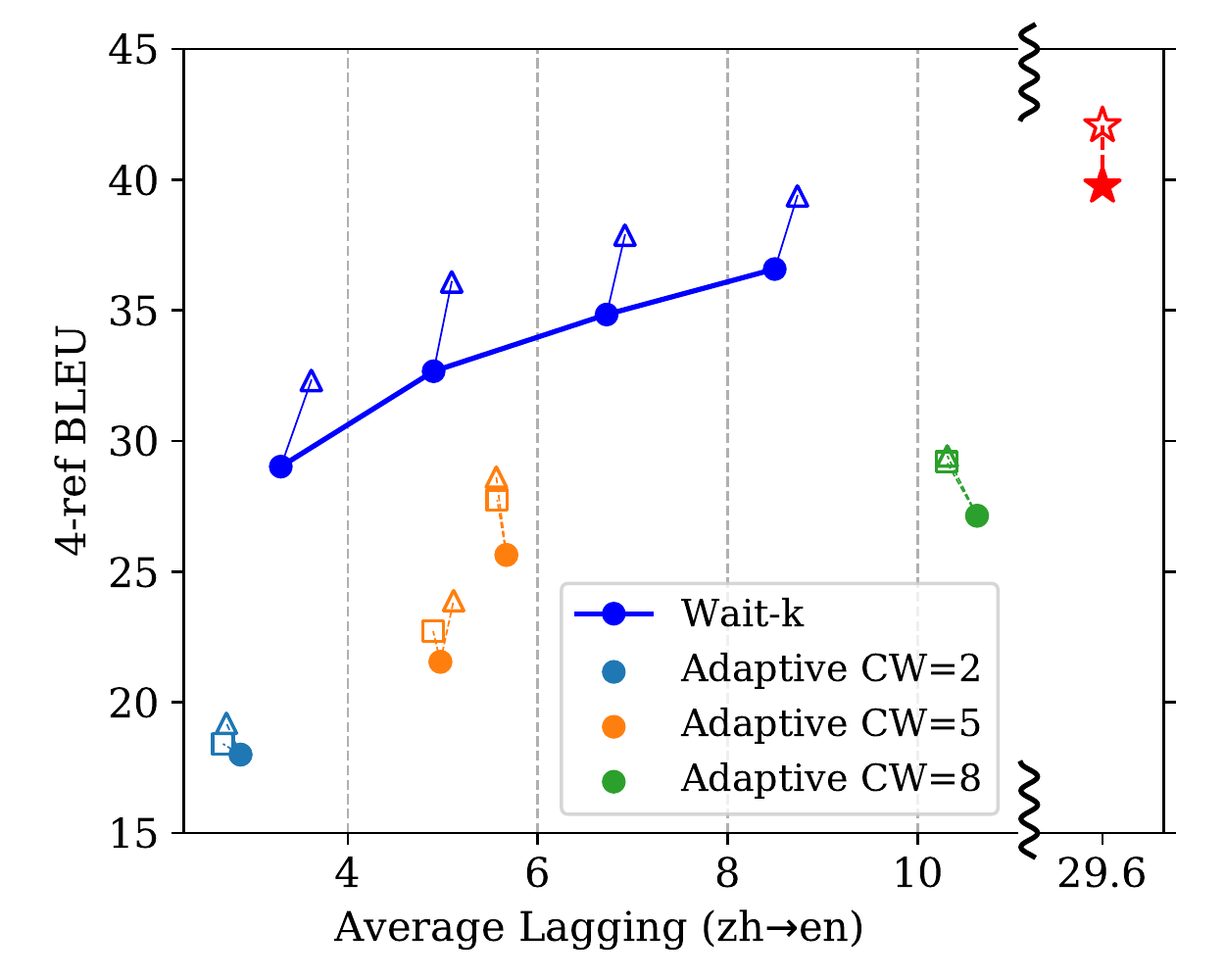} & \qquad
  \centering
  \includegraphics[width=6.9cm]{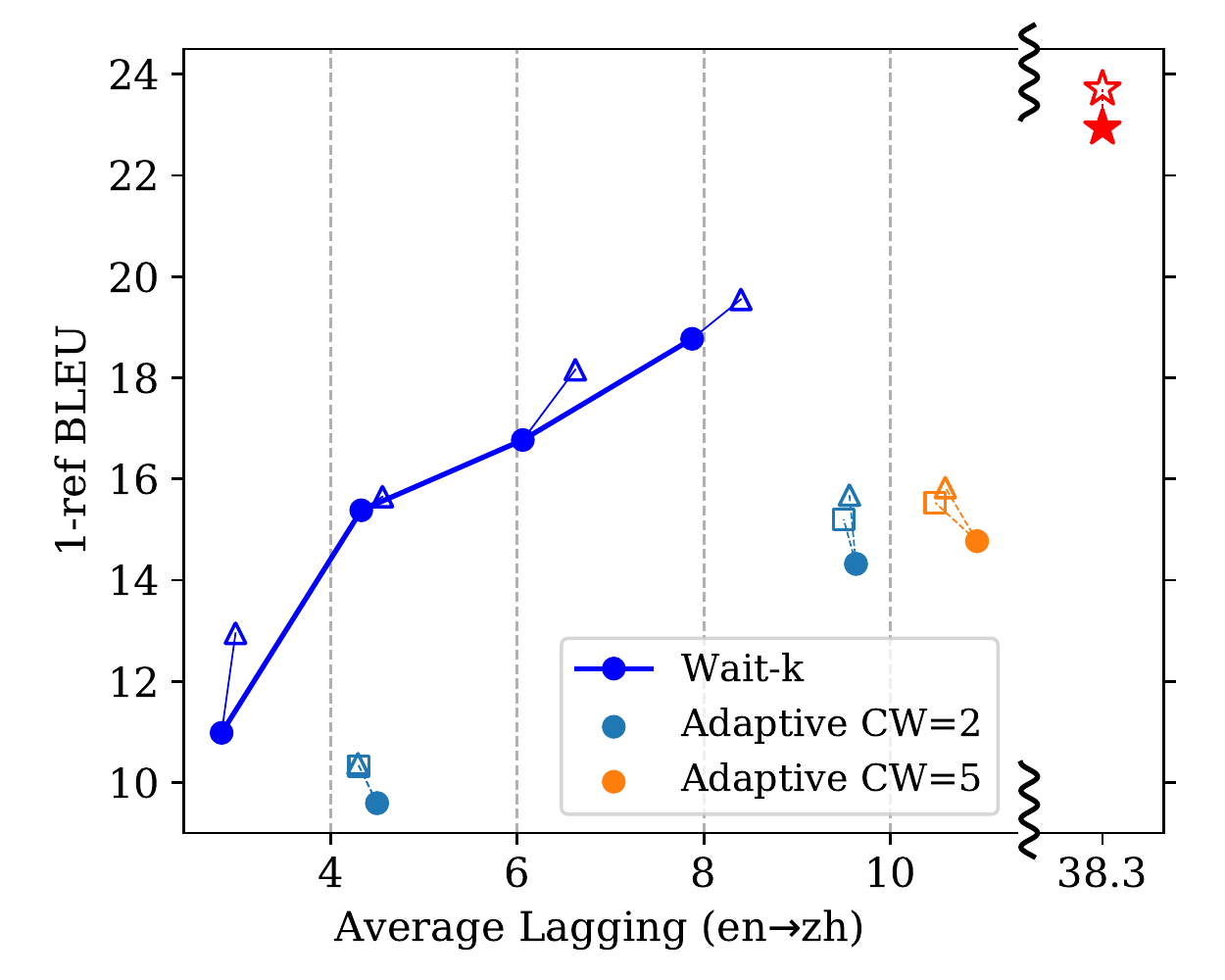}
\end{tabular}\\[-0.4cm]
\captionof{figure}{
  BLEU against AL
  using
  adaptive policy (compared with wait-$k$ models)
   with
  different beam search methods.
   \textcolor{colorC0}{$\square$} \textcolor{colorC1}{$\square$} \textcolor{green}{$\square$}: conventional beam search in chunk of consecutive write \cite{gu+:2017}.
   \textcolor{colorC0}{$\triangle$} \textcolor{colorC1}{$\triangle$} \textcolor{green}{$\triangle$}: speculative beam search.
  \textcolor{red}{$\bigstar$\ding{73}}:full-sentence baseline (greedy and beam-search).
}
\label{fig:zh2en_rl}
\end{figure*}

We evaluate our work on Chinese$\leftrightarrow$English simultaneous translation tasks. 
For the training data, we use the   
NIST corpus for Chinese$\leftrightarrow$English (2M sentence pairs).
We first apply BPE~\cite{sennrich+:2015} on all texts in order 
to reduce the vocabulary sizes. 
For Chinese$\leftrightarrow$English evaluation,
we use NIST 2006 and 
NIST 2008 as our dev and test sets 
with 4 English references.
For English$\to$Chinese, we use the second among the four English references as the source text.

We re-implement wait-$k$ model \cite{ma+:2018}, 
test-time wait-$k$ model \cite{dalvi+:2018}
and adaptive policy \cite{gu+:2017} 
based on PyTorch-based OpenNMT \cite{klein+:2017}.
To reach state-of-the-art performance, we use Transformer based
wait-$k$ model and also use Transformer based
pre-trained full sentence model for learning adaptive policy.
The architecture of Transformer is the same as the base model from 
the original paper \cite{vaswani+:2017}.
We use Average Lagging (AL) \cite{ma+:2018} as the latency metrics.
AL measures the number of words
delay for translating a given source sentence.

\begin{table}[!htb]\centering
\small
\begin{tabular}{|l|c|c|c|c|c|c|c|}\hline
\diagbox{\!$b$}{\! $w$} & 0 & 1     & 2     & 3     & 4      & 5 \\\hline
 1 & 34.57 & -  & - & - & -  & - \\\hline
 3 & -   & +1.3  & +1.8 & +1.2 & +2.0  & +1.7 \\\hline
 5 & -   & +1.6  & +1.9 & +1.3 & +1.5  & +1.3 \\\hline
 7 & -   & +1.5  & +2.0 & +1.0 & +1.6  & +1.4 \\\hline
 10 &-   & +1.4  & +2.2 & +1.4 & +1.5  & +1.7 \\\hline
\end{tabular}
\caption{Zh$\to$En wait-$1$ model BLEU improvement of
SBS against greedy search ($b=1$, $w=0$) on dev-set. When $w \ge 5$ 
the performance of SBS becomes stable.}
\label{tab:w1}
\vspace{-0.1cm}
\end{table}


\begin{figure*}[!ht]
\resizebox{\textwidth}{!}{%
\setlength{\tabcolsep}{1pt}
\centering
\begin{tabu}{ c c | l l l l l l l l l l  }
\rowfont{\small}
& & \chn{sh\`{\i}h\'ang} &\chn{n\v{\i}} & \chn{j\v{\i}anm\v{\i}an} & \chn{z\`ui} &
\chn{qi\'ong} & \chn{g\'uoji\=a} & \chn{zh\`aiw\`u} &   \\

 &  & 世行 & 拟 & 减免 &最&穷&国家&债务 & \\ 

 \rowfont{\small}
Gloss &  & world bank & plan to & remit \& reduce & most  & poor & country & debt &  \\  
 \hline
\multirow{2}{*}{$k$=1$^\dagger$} &  Greedy & & world & bank& to & reduce & poverty & - & stricken countries &  \\
  & SBS &  & world& bank & to &  exemp- & t & po- & or- est countries from debt &  \\ \hline
\multirow{2}{*}{$k$=1$^\ddagger$}  & Greedy &  & world & bank & to & reduce & or & exemp- & t  debt  of po- or- est countries   \\[0.05cm]
  & SBS &  & world & bank  & inten- & ds & to & reduce & or  exemp- t debt of po- or- est countries   \\[0.05cm]
 \hline
$k$=$\infty^\ast$ &  & \multicolumn{7}{l}{} & world bank plans to remit and reduce debts of po- or- est countries  \\[-0.3cm] 
\end{tabu} 
}
\caption{Chinese-to-English example on dev set. 
  $^\dagger$: test-time wait-$k$; $^\ddagger$: \waitk. $^\ast$: full-sentence beam search.
  \vspace{-9pt}
}
\label{fig:example}
\end{figure*}

\subsection{Performance on Wait-$k$ Policy}

We perform experiments on validation set using speculative beam search (SBS)
with beam sizes $b \in \{3, 5, 7, 10\}$
and speculative window sizes $w \in \{1, 2, 3, 4, 5\}$.
Table~\ref{tab:w1} shows the BLEU score
of different $b$ and $w$ over
wait-$1$ model.
Compared with greedy decoding, SBS
improves at least 1.0 BLEU score in all cases and
achieves best performance by $b= 10, w=2$.
We search the best $b$ and $w$ for each model on dev-set
and apply them on test-set in the following experiments.

Fig.~\ref{fig:zh2en_al} shows
the performance of conventional greedy decoding,
trivial tail beam search (only after source sentence is finished) and SBS
on test set on Chinese$\leftrightarrow$English tasks.
SBS largely boost test-time wait-$k$ models with 
slightly worse latency (especially in English$\to$Chinese
because they tend to generate longer sentences).
Wait-$k$ models also benefit from speculation (especially in
Chinese$\to$English).

Fig.~\ref{fig:example} shows a running example of
 greedy and SBS output of
both wait-$k$ and test-time wait-$k$ models.
SBS on test-time wait-$k$ generates much better
outputs than the greedy search, which misses 
some essential information.
Wait-$k$ models with speculation correctly translates ``拟'' into ``intends to''
instead of ``to'' in greedy output.

\begin{figure*}[!htb]
\begin{tabular}{cc}
\centering
  \includegraphics[width=6.5cm]{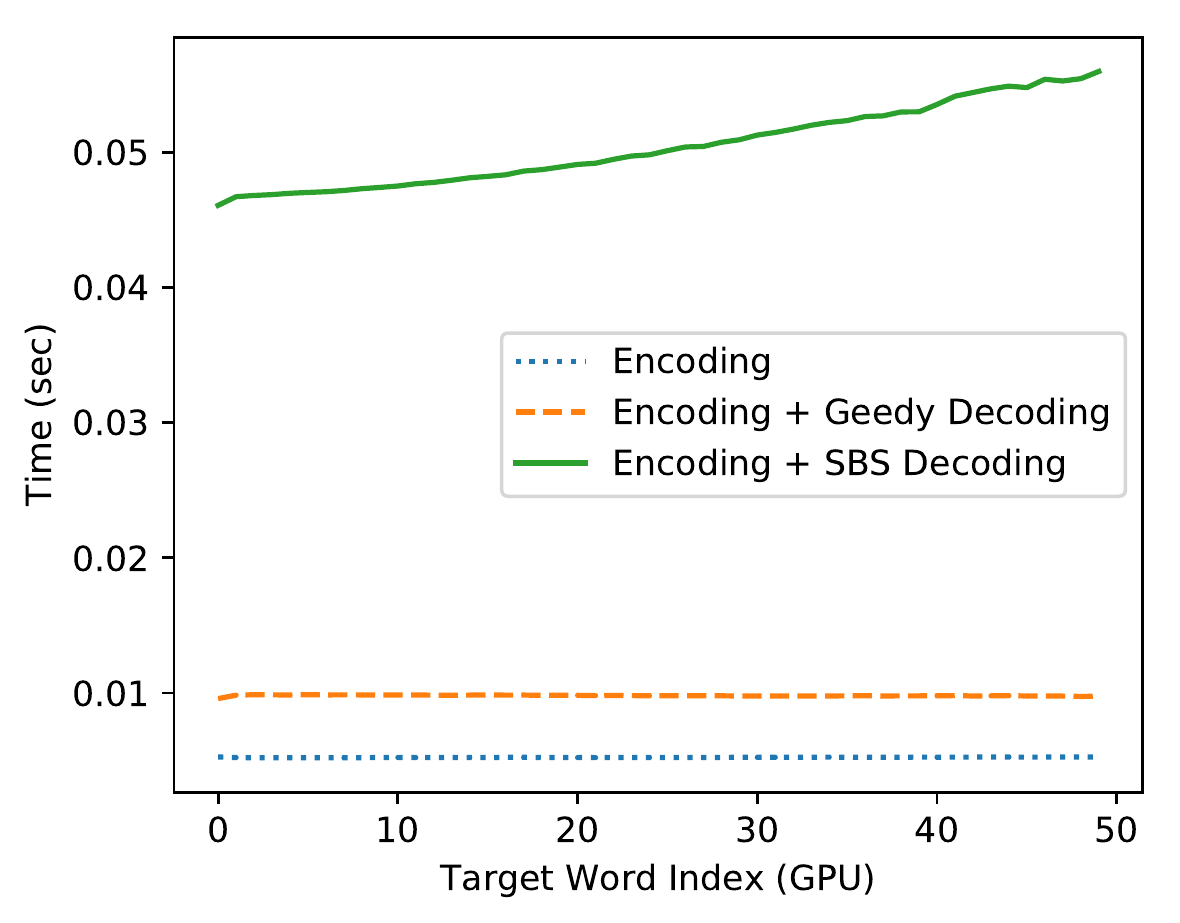} & \centering \qquad \qquad
  \includegraphics[width=6.5cm]{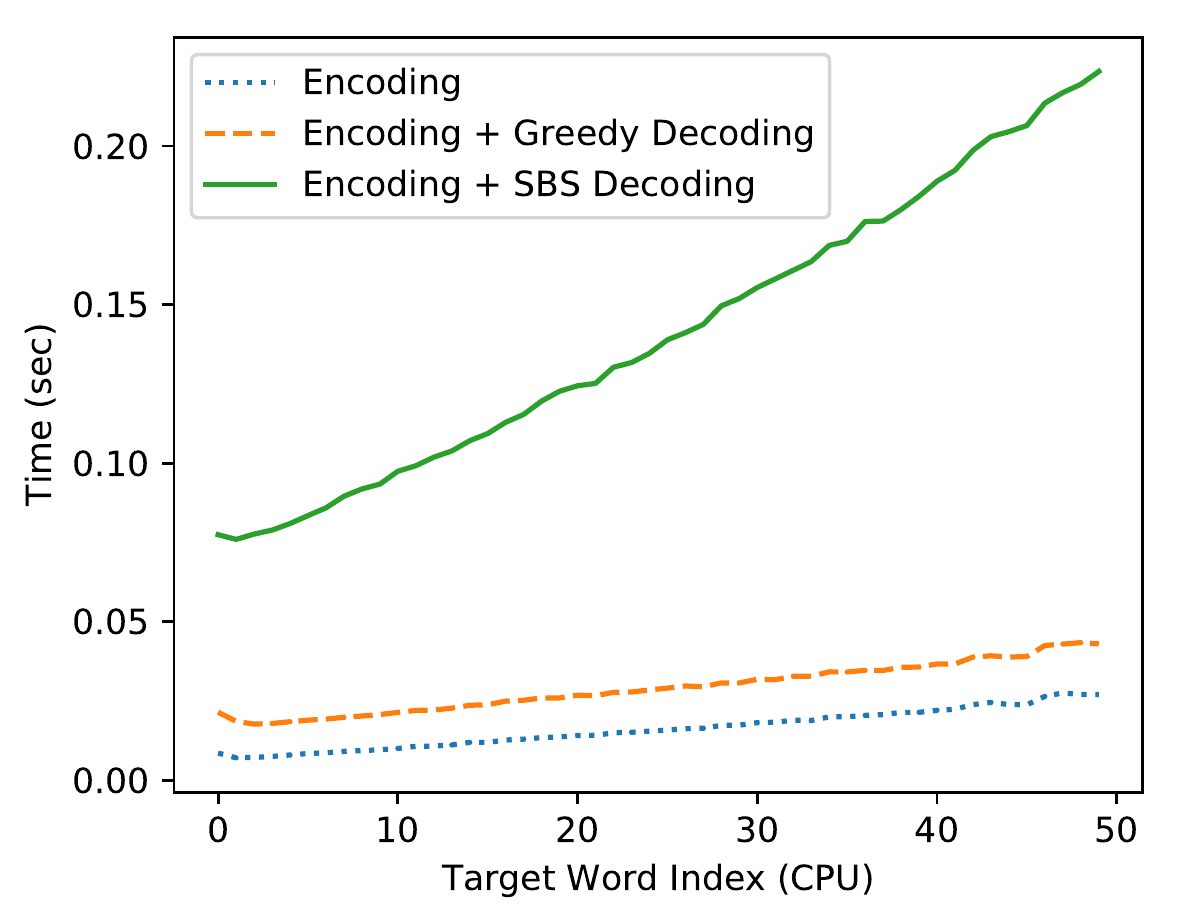}
\end{tabular}\\[-0.4cm]
\captionof{figure}{
  Average time of words with different indices ($\leq 50$)
   on zh$\to$en wait-3 model.
  Results in left figure are measured on GPU
  while results in right figure results are measured on CPU.
  The SBS results adopt $w=2, b=5$.
}
\label{fig:zh2en_time}
\vspace{-10pt}
\end{figure*}

\begin{figure}[!htb]
\begin{tabular}{cc}
\centering
  \includegraphics[width=5.5cm]{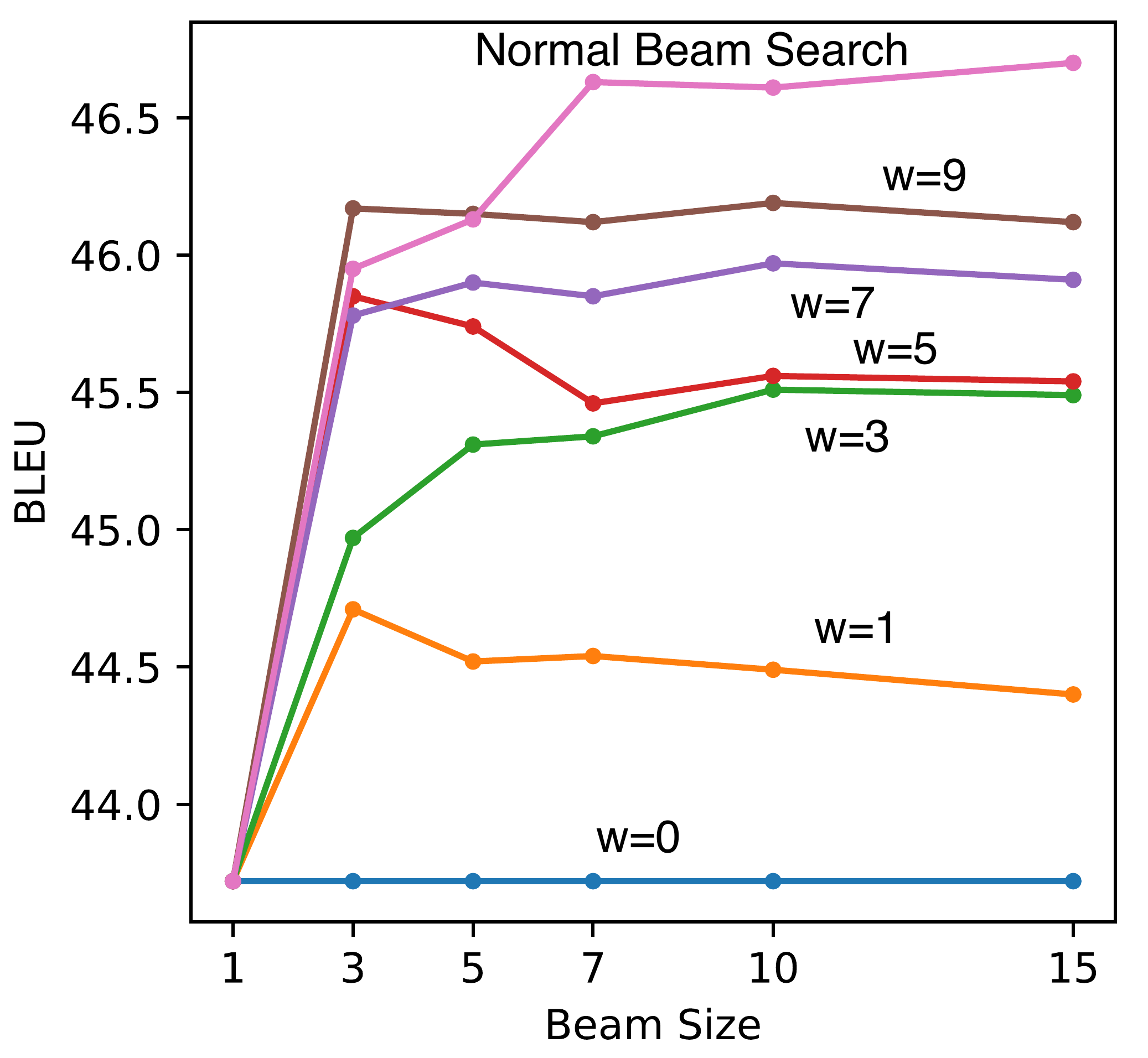} & \centering
\end{tabular}\\[-0.4cm]
\captionof{figure}{
  BLEU of SBS over full sentence translation 
  with different window sizes $w$ and beam sizes $b$.
  Window size $w=0$ stands for greedy decoding.
  The top line stands for results using normal beam search.
}
\label{fig:zh2en_full}
\vspace{-10pt}
\end{figure}

\subsection{Performance on Adaptive Policy}
\vspace{-4pt}

Fig.~\ref{fig:zh2en_rl} shows the performance of proposed
SBS on adaptive policies.
We train adaptive policies using the combination
of  Consecutive Wait (CW $\in \{2, 5, 8\}$~\cite{gu+:2017}) and partial-BLEU as reward in
reinforcement learning.
We vary beam size $b \in \{5, 10\}$ in both
chunk-based beam search \cite{gu+:2017} and our  
SBS with speculative
window size $w \in \{2, 4\}$.
Our proposed beam search achieves better results
in most cases.


\vspace{-4pt}
\subsection{Running Time Analysis}
\vspace{-4pt}

Fig.~\ref{fig:zh2en_time} shows the average time for generating 
words with different target word indices on a GeForce GTX TITAN-X GPU and an Intel Core i7 2.8 GHz CPU.
According to~\namecite{ma+:2018}, wait-$k$ models use
bi-directional Transformer as the encoder,
thus the time complexity of incrementally encoding one more source word
is $O(m^2)$ where $m$ is the source sentence length.
This is the reason why it takes more time to encode words with
larger index especially using CPU.
It is generally accepted that Mandarin speech is about 120--150 syllables per minute, and
in our corpus each token (after BPE) has on average 1.5 Chinese syllables
(which is 1.5 characters since each Chinese character is monosyllabic),
thus in the simultaneous Chinese-to-English speech-to-text translation scenario,
the decoder receives a source token every 0.6--0.75 seconds
which is much slower than our decoding speed (less than 0.25 seconds per token) even on a laptop CPU.
Based on these statistics, our proposed speculative beam search
algorithm can be used in
real simultaneous translation.

\vspace{-4pt}
\subsection{Performance on Full Sentence MT}
\vspace{-4pt}

We analyze the performance of  speculative
beam search on full-sentence translation
(see Fig.~\ref{fig:zh2en_full}).
By only performing beam search on a sliding speculative
window, the proposed algorithm achieves much better BLEU scores
compared with greedy decoding ($w=0$)
and even outperforms conventional beam search
when $w=9$, $b=3$.
Please note that the space complexity of this algorithm is
$O((m+n+wb)d)$.{\footnote{Here 
$n$ is the length of target sentence and 
$d$ is the representation dimension.}}
This is better than conventional beam search whose space complexity is
 $O((m+nb)d)$ when $w \ll n$.

\section{Conclusions and Future Work}
\vspace{-4pt}

We have proposed speculative beam search 
for simultaneous translation.
Experiments 
on three approaches to simultaneous translation 
demonstrate effectiveness of our method.
This algorithm has the potential
 in other incremental tasks such as streaming ASR and incremental
TTS.
\section*{Acknowledgments}
         {
           \vspace{-0.2cm}
We thank Kaibo Liu for his AL script\footnote{\href{https://github.com/SimulTrans-demo/STACL}{\scriptsize \tt https://github.com/SimulTrans-demo/STACL}}
and help in training wait-$k$ models, and the anonymous reviewers for suggestions. 
}

\balance

\bibliographystyle{acl_natbib}
\bibliography{main}


\clearpage

\appendix

\section{Supplemental Material}


We also evaluate our work using Consecutive Wait (CW) 
 as latency metric, which measures the average lengths
 of consecutive wait segments,
and perform experiments on
German$\leftrightarrow$English
corpora available from WMT15\footnote{\href{http://www.statmt.org/wmt15/translation-task.html}{\scriptsize \tt http://www.statmt.org/wmt15/translation-task.html}}.
We use newstest-2013 as dev-set and newstest-2015 as test-set.\footnote{The German$\leftrightarrow$English results are slightly different from those in \namecite{ma+:2018} because 
of different decoding settings.
We do not allow that the decoder stops earlier than the finish of source sentence while it is allowed in German$\leftrightarrow$English experiments of \namecite{ma+:2018}.
This makes our generated sentences longer and further results in worse AL
compared with the results in \namecite{ma+:2018}.}

\begin{figure}[!hb]
\begin{tabular}{cc}
\centering
  \includegraphics[width=7.cm]{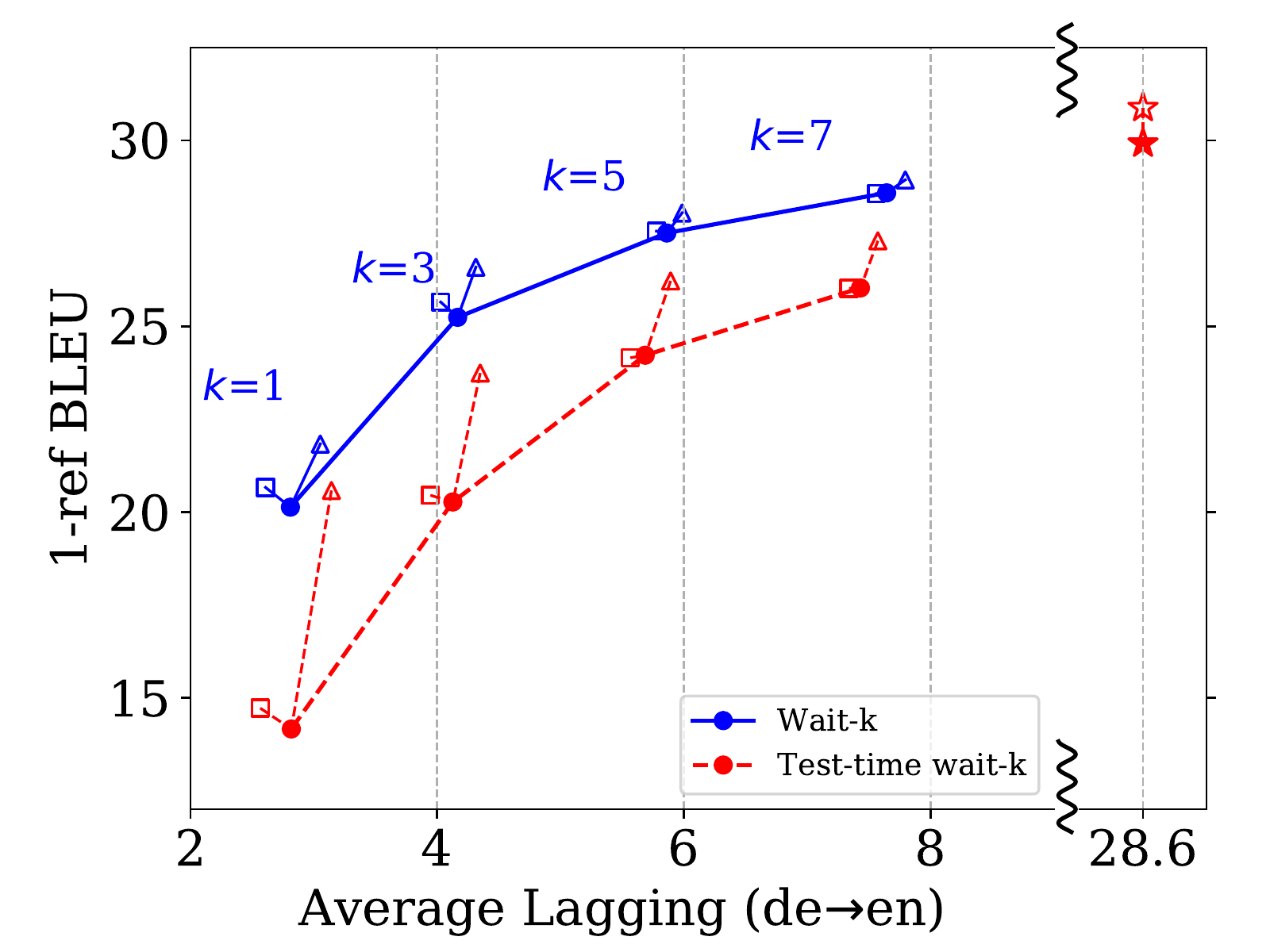} & \\
 \centering
  \includegraphics[width=7.cm]{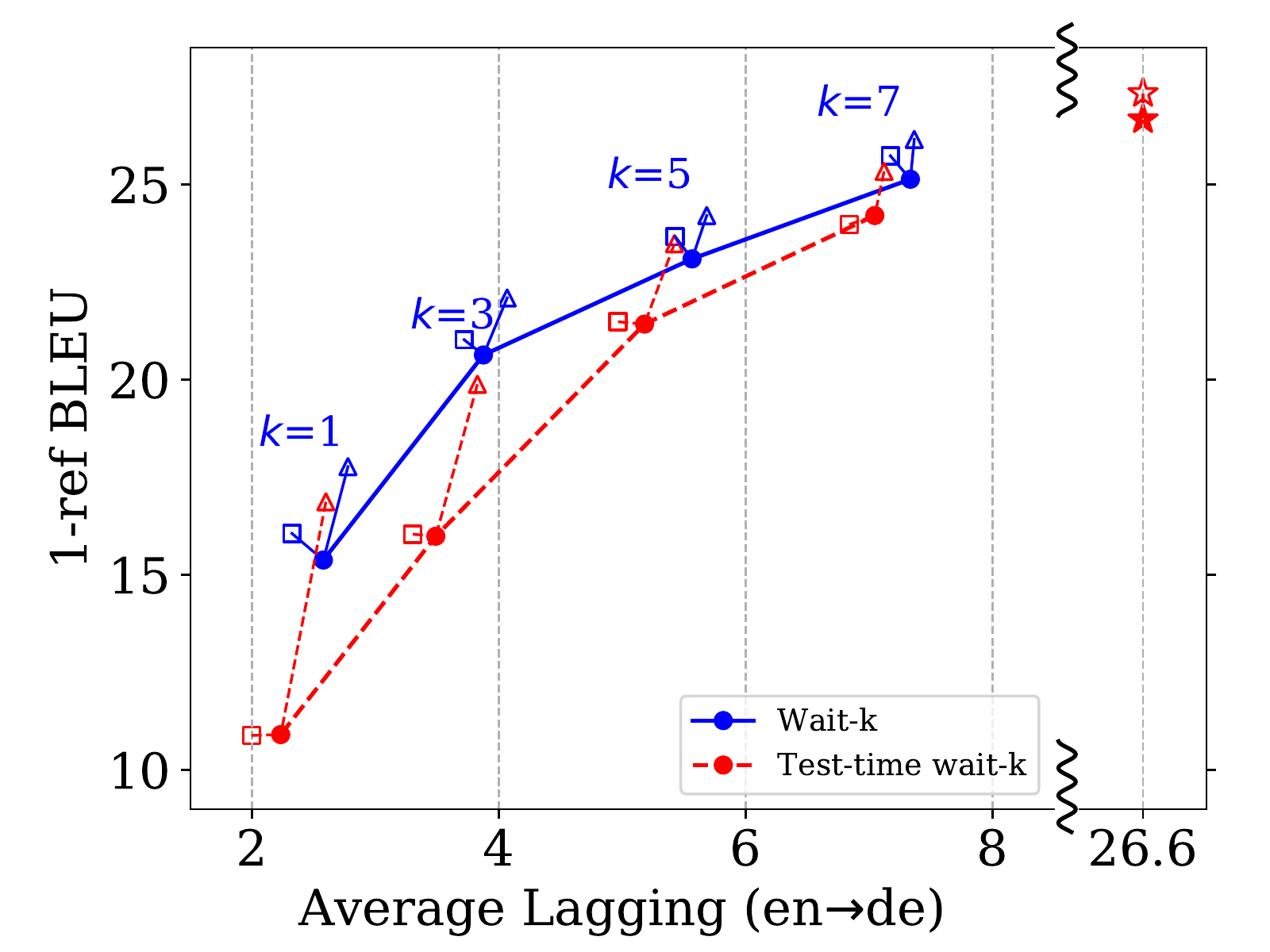}
\end{tabular}\\[-0.3cm]
\captionof{figure}{
  Translation quality against AL on English$\leftrightarrow$German simultaneous translation using wait-$k$ model.
  \textcolor{red}{$\square$} \textcolor{blue}{$\square$}: conventional beam search only on target tail.
  \textcolor{red}{$\triangle$} \textcolor{blue}{$\triangle$}: speculative beam search.
  \textcolor{red}{$\bigstar$\ding{73}}:full-sentence (greedy and beam-search).
}
\label{fig:de2en_ap}
\vspace{-0.7cm}
\end{figure}

\begin{figure}[!hb]

\begin{tabular}{cc}
\centering
  \includegraphics[width=7.cm]{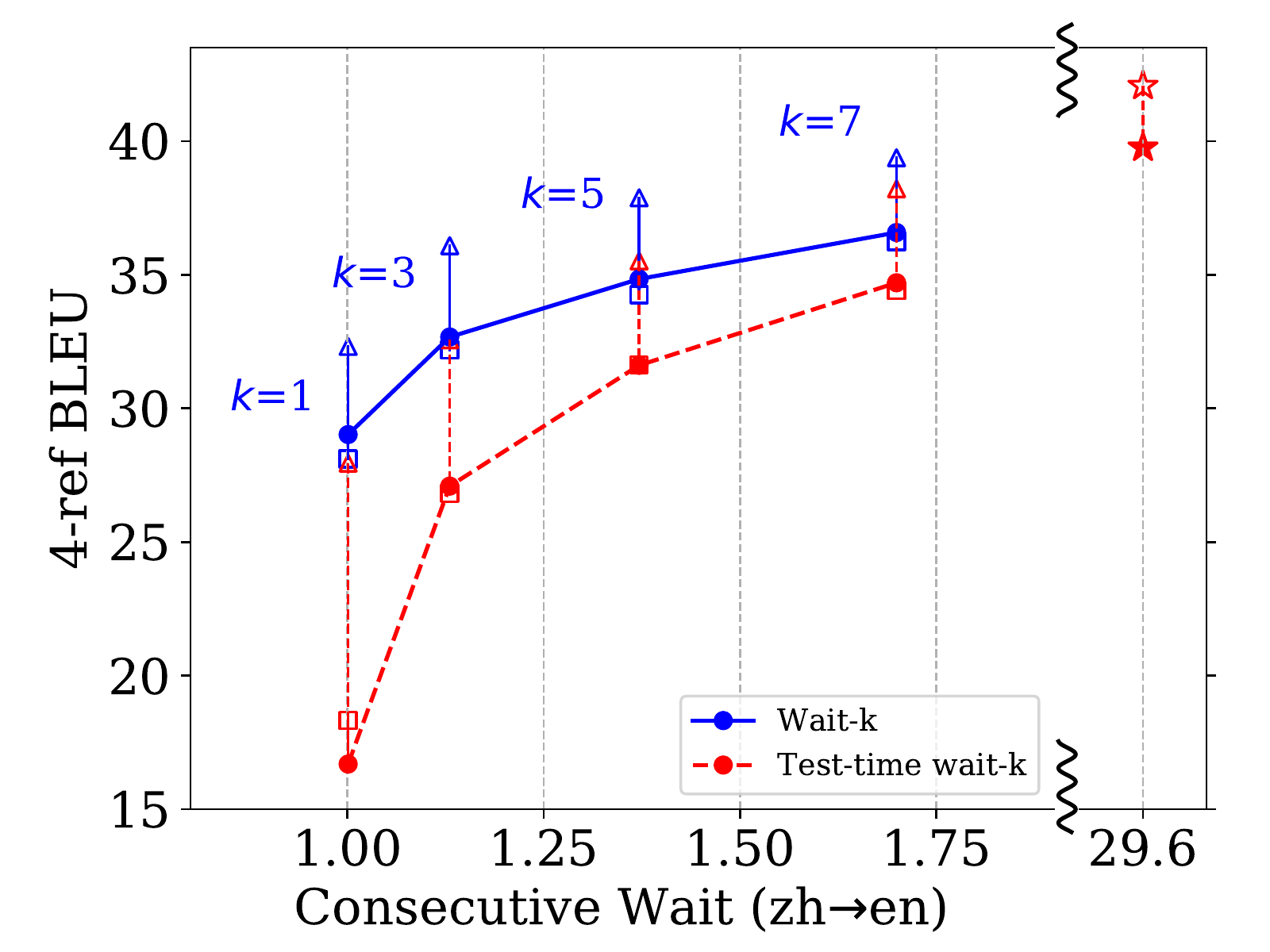} & \\ \centering
  \includegraphics[width=7.cm]{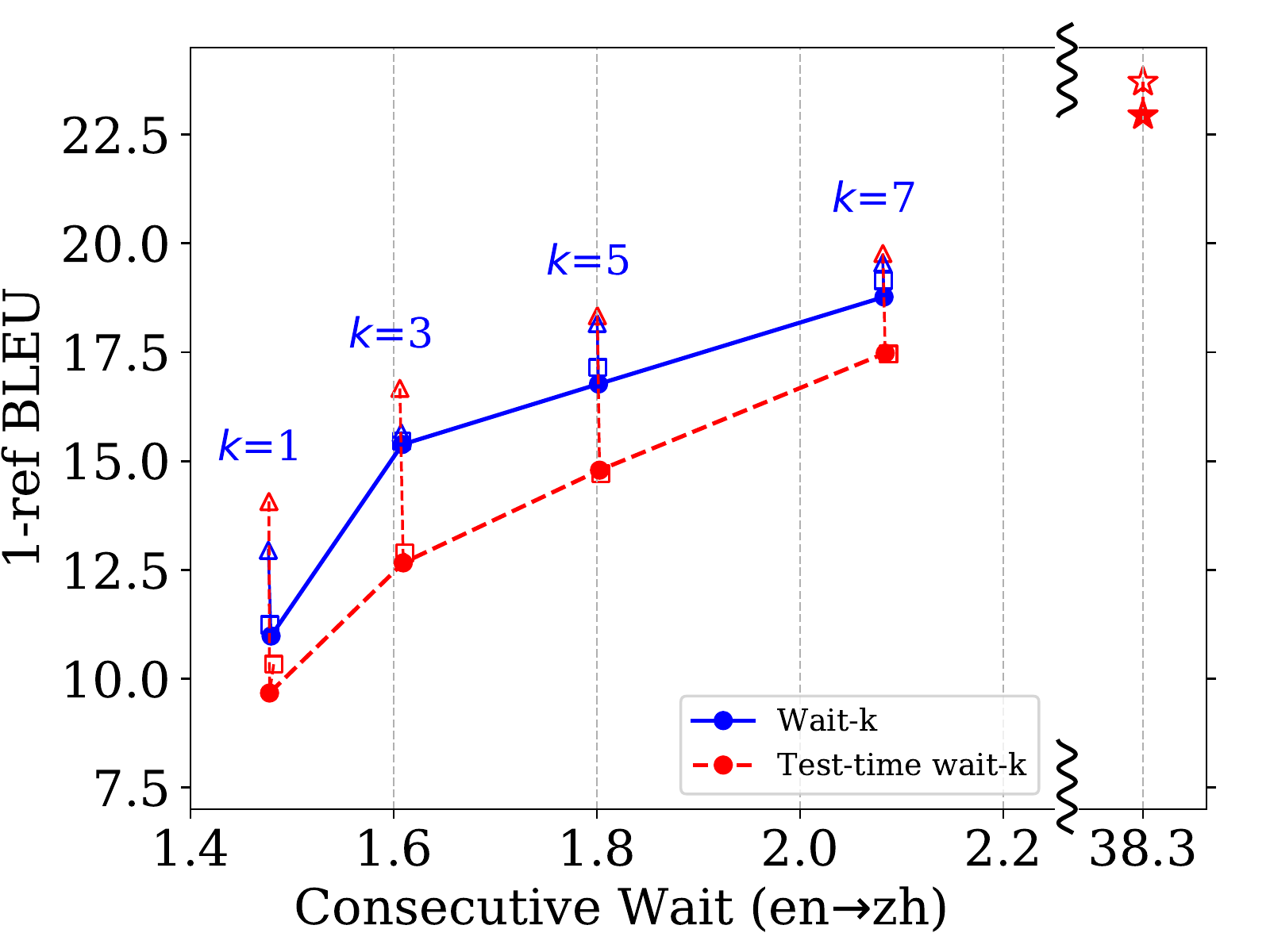}
\end{tabular}\\[-0.3cm]
\captionof{figure}{
  Translation quality against CW on Chinese$\leftrightarrow$English simultaneous translation using wait-$k$ model.
}
\label{fig:zh2en_cw}

\end{figure}

\begin{figure}[!hb]
\begin{tabular}{cc}
\centering
  \includegraphics[width=7.cm]{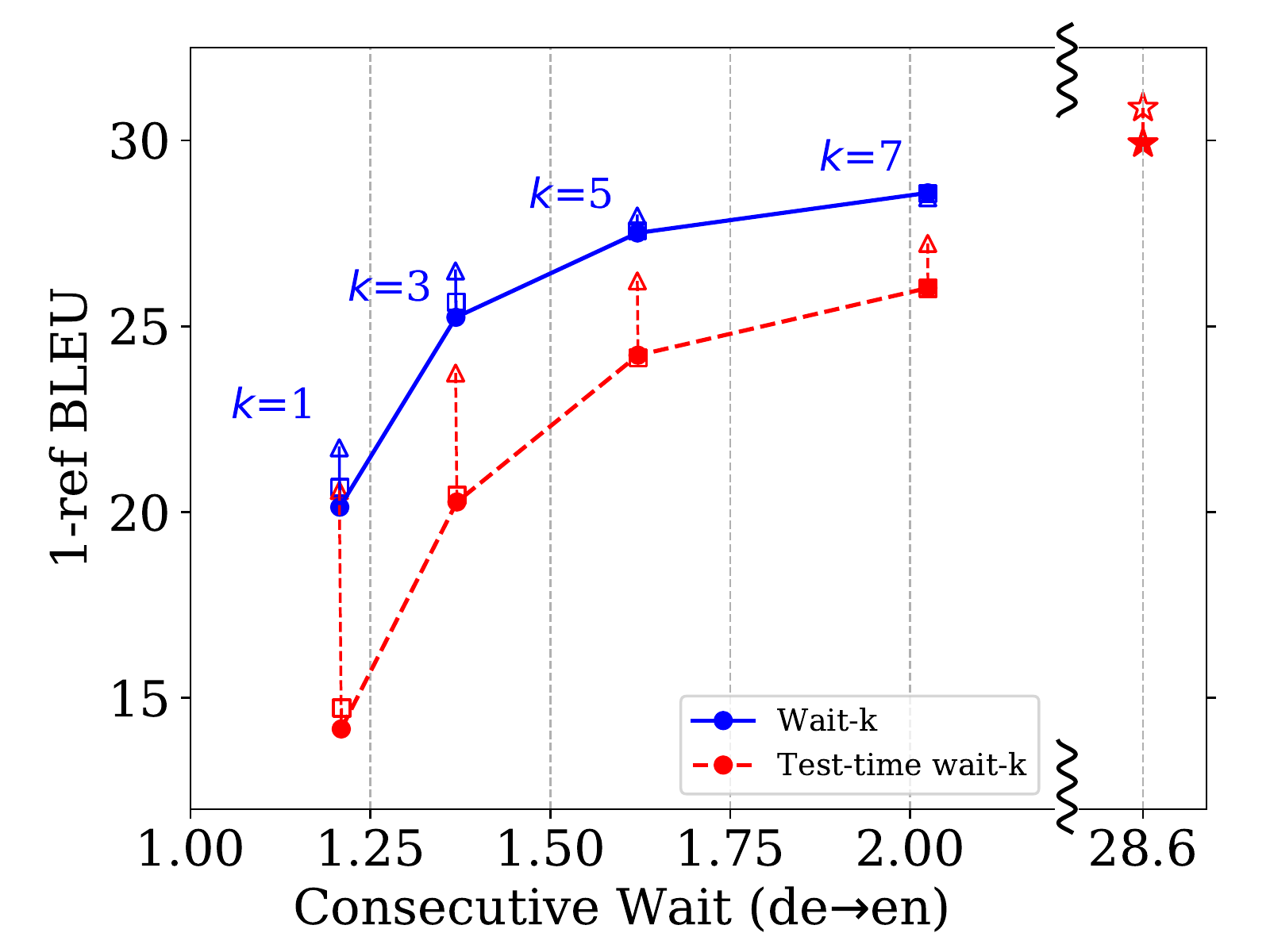} & \\ \centering
  \includegraphics[width=7.cm]{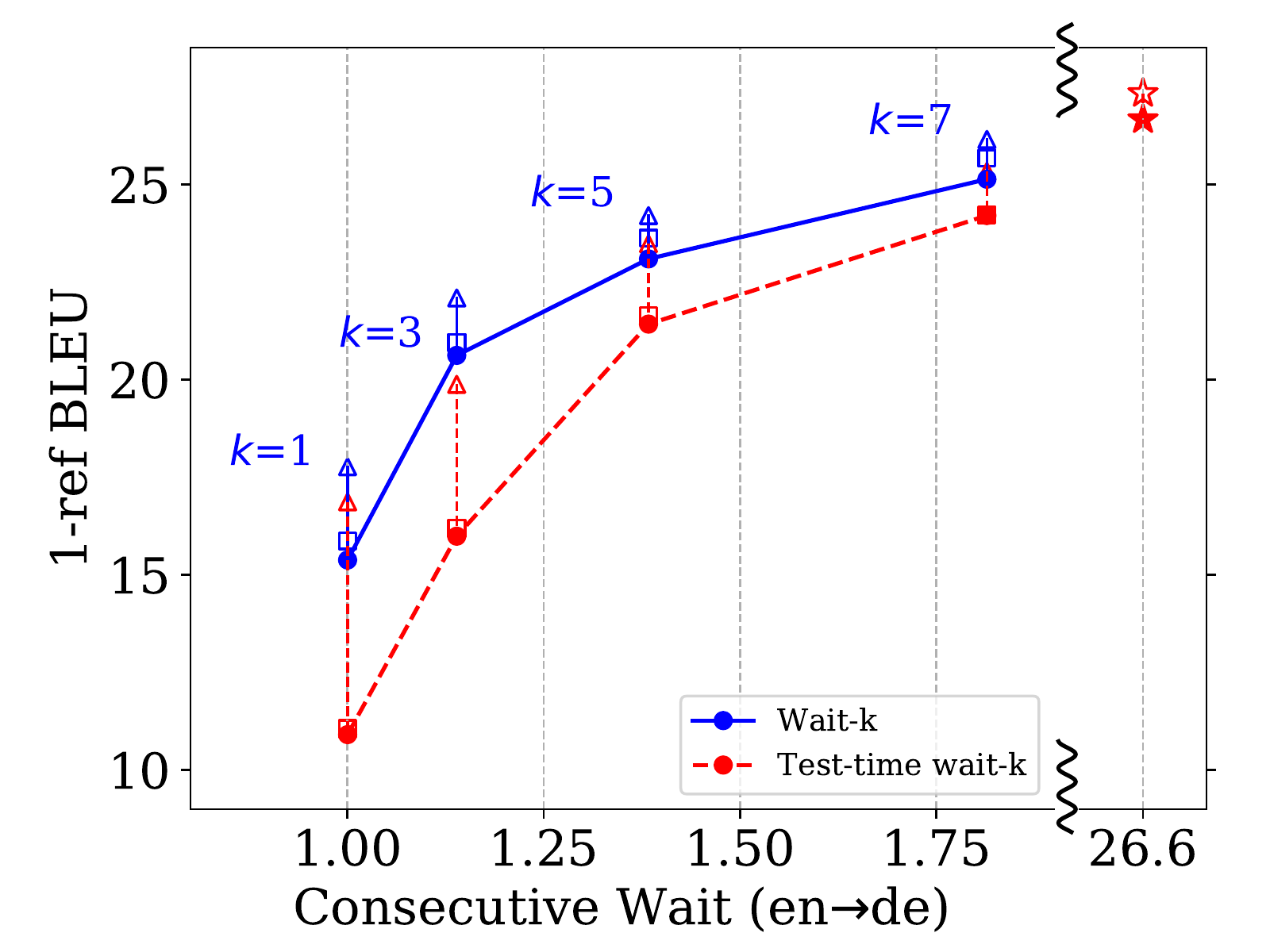}
\end{tabular}\\[-0.3cm]
\captionof{figure}{
  Translation quality against CW on English$\leftrightarrow$German simultaneous translation using wait-$k$ model.
}
\label{fig:de2en_cw}
\end{figure}

Fig.~\ref{fig:de2en_ap} show the translation quality
on German$\leftrightarrow$English against AL of
different decoding methods.
Consistent to the results of Chinese$\leftrightarrow$English,
our proposed speculative beam search gain large performance
boost especially on test-time wait-$k$.
Fig.~\ref{fig:zh2en_cw} and Fig.~\ref{fig:de2en_cw}
use CW as latency metrics. Since both the wait-$k$ 
and test-time wait-$k$ models use the same
fixed policy, the CW latencies of the same $k$ are identical.



\end{CJK}
\end{document}